# DARTH VECDOR: AN OPEN-SOURCE SYSTEM FOR GENERATING KNOWLEDGE GRAPHS THROUGH LARGE LANGUAGE MODEL QUERIES

**Author**: Jonathan A. Handler, MD (1, 2)

**Author Affiliations**:

1) Keylog Solutions LLC, Northbrook, IL, USA. jhandler@gmail.com
2) Clinical Intelligence and Advanced Data Lab, OSF HealthCare, Peoria, IL, USA. jonathan.a.handler@osfhealthcare.org

## ABSTRACT

Many large language models (LLMs) are trained on a massive body of knowledge present on the Internet. Darth Vecdor (DV) was designed to extract this knowledge into a structured, terminology-mapped, SQL database ("knowledge base" or "knowledge graph"). Knowledge graphs may be useful in many domains, including healthcare. Although one might query an LLM directly rather than a SQL-based knowledge graph, concerns such as cost, speed, safety, and confidence may arise, especially in high-volume operations. These may be mitigated when the information is pre-extracted from the LLM and becomes query-able through a standard database. However, the author found the need to address several issues. These included erroneous, off-topic, free-text, overly general, and inconsistent LLM responses, as well as allowing for multi-element responses. DV was built with features intended to mitigate these issues. To facilitate ease of use, and to allow for prompt engineering by those with domain expertise but little technical background, DV provides a simple, browser-based graphical user interface. DV has been released as free, open-source, extensible software, on an "as is" basis, without warranties or conditions of any kind, either express or implied. Users need to be cognizant of the potential risks and benefits of using DV and its outputs, and users are responsible for ensuring any use is safe and effective. DV should be assumed to have bugs, potentially very serious ones. However, the author hopes that appropriate use of current and future versions of DV and its outputs can help improve healthcare.

## INTRODUCTION

Large language models (LLMs) have already had a significant impact in healthcare, and many more uses are reported in development and in planned implementation. Since LLMs are trained on huge volumes of data, LLMs are encoded with a significant swath of the knowledge present on the Internet and possibly other sources. Therefore, the author hypothesized that LLMs can be used to as a source to populate knowledge graphs in a database (or "knowledge base") for various uses. For example, a knowledge graph of medications used to treat diseases might be used as a part of a research effort in which a database that includes the knowledge graph along with patient data is queried to find which patients have potentially untreated diseases (i.e., no medication has been prescribed that treats that disease).

Querying a knowledge graph previously created through LLM queries, rather than just querying an LLM directly as needed, may have several potential advantages:

1. **Cheaper**
    a. **Lower compute costs**: In some cases, the cost of computation to query a knowledge graph may be dramatically lower than querying an LLM.
    b. **Lower hardware costs and complexity**: If the LLM that would have been used would be run on institutionally controlled servers, the costs, complexity, and management burden of the hardware stack required to achieve rapid responses may be prohibitive for many. The hardware costs and complexity needed for querying a knowledge graph in a database rather may be much lower in many cases than those needed to support many LLMs.
2. **Faster**
    a. **Faster query speed**: In many cases, querying a knowledge graph (e.g., via a vector database) may be orders of magnitude faster than querying an LLM.
    b. **Facilitation of development and implementation**: The people, processes, and technologies for building and implementing systems built on a knowledge graph (perhaps especially if implemented through a SQL database) may be more well-developed and readily available than systems using LLMs.
3. **Safer**
    a. **Reduction of privacy and confidentiality risks**: If the LLM that would have been used is a third party's commercial service, querying a knowledge graph running on institutionally controlled servers instead may reduce the risks of submitting potentially sensitive data to a commercial third-party service.
    b. **Reduction of many business risks**: If the LLM is controlled by a third party, then using a knowledge graph in operational use rather than directly querying the LLM may reduce many business risks such as the third party deprecating the product, increasing pricing, or modifying functionality.
4. **Surer**
    a. **More explainable responses**: LLMs are often considered "black boxes" since the actual logic for producing a given output commonly cannot be provided in a format meaningful to humans. Although an LLM's population of the knowledge graph may be considered "black box," the downstream usage of that knowledge graph can often be more explainable since the graph can be visualized, and the queries can be explained.
    b. **More consistent responses**: LLMs may give inconsistent responses when repeatedly given the same prompt, especially if they are allowed to use "creativity" when responding. Therefore, an LLM may not populate the graph deterministically (i.e., if the graph were repopulated, some or all content might differ). However, once populated, the knowledge graph remains stable and will return consistent results to any deterministic query for as long as the knowledge graph remains unmodified.
    c. **More easily validated content**: Since the knowledge base content can be queried and viewed in ways that return perfectly consistent results, it may be easier to validate the content and/or use cases using that content.

To successfully use LLMs for this purpose, common challenges and failure modes must be identified and mitigated. During this work, this researcher identified the following key set of items to be mitigated, albeit non-exhaustive and at least some of which are widely known LLM issues while others may be much less so, perhaps because they more directly relate to the creation of knowledge graphs in general, and healthcare knowledge graphs specifically:

- **Erroneous or off-topic responses**: for example, responding that anticoagulation is a common treatment for hemorrhage (since anticoagulation is generally a cause of, not a treatment for hemorrhage); or responding with "radiation therapy" when asking for a list of medication names that treat cancer (since radiation would not normally be considered a medication).
- **Free-text responses**: Coding systems and medical terminologies often serve as the "lingua franca" of healthcare software systems and associated databases. Many uses of a knowledge base with health system data require the knowledge base's content to be encoded into a medical terminology.  However, during this work the LLM under study seemed far more likely to hallucinate when asked to return results as structured codes (e.g., ICD-10) than as free text.
- **Overly general responses**: Even when asked to return very specific responses and to avoid a broad category as a response (such as "antibiotics"), the experiences would include a broad category as a response or as part of a response more often than desired. Knowledge graphs having broad category objects may have less utility for many use cases.
- **Inconsistent responses**: Even when asked to return a response limited to a specific list of text responses, occasionally an LLM would return a slightly different response. When the need was for the response to contain only a member of the list, even slightly different responses were problematic as the correct response would have to be inferred.
- **Multi-element responses**: Very often, multi-component responses are requested or produced, and these then need to be parsed into the individual component elements. For example, lists need to be parsed into their component elements.

The author hypothesized that methods (previously reported or newly developed) could at least partially mitigate these issues, and that a platform utilizing these and other methods to facilitate the creation and use of knowledge graphs populated by LLMs would have benefit. This work describes some aspects of "Darth Vecdor" and its output. Darth Vecdor (DV) is free, open-source software developed by the author to generate knowledge graphs from various sources, including queries of other knowledge graphs, as well as queries of an LLM.[1] The focus of this work is based on a knowledge base populated by DV using queries of a base (not fine-tuned) LLM.

## SOFTWARE DESCRIPTION

### ABOUT DARTH VECDOR

Darth Vecdor [1] is a software platform written in Python v3.10.2 (free, open-source) and implemented using PostgreSQL v15.10 as the underlying database (free, open-source) with the pgVector extension v0.8.0 (free, open-source) to enable vector use in the database.[2], [3], [4] The source code is available at https://github.com/jonhandlermd/darth_vecdor. While versions may have been updated at some point during development, these were the versions as of the time of the writing of this paragraph, and likely the versions used during the majority of development. DV also uses many other libraries, with heavy use of SQL Alchemy for its database interactions.[5] DV was designed to be extensible, and users can build additional modules to allow DV to work with different embedding models and different LLMs. DV was primarily configured during development to query OpenAI's "gpt-4o-mini" LLM model.[6] During development, most embeddings were generated using the NeuML pubmedbert-base-embeddings model (abbreviated here as "PMBert"), which is reported as a MSR BiomedBERT (abstracts + full text) base model fine-tuned using the SentenceTransformers library.[7], [8], [9]

In brief, embeddings are vectors intended to represent the semantic content ("meaning") of text. SentenceTransformers were selected for use as they were designed to perform embeddings considering the meaning of words within the context of a sentence (or phrase), which seemed appropriate for terminologies often having multi-word entries. PMBert was selected for this work as it has been reported to have "state-of-the-art performance on many biomedical NLP tasks."[7] OpenAI's ChatGPT gpt-4o-mini LLM model ("gpt-4o-mini") was chosen for this work due to its relatively low cost, speed, and reported high-level of performance,[6] along with a robust API and its ability to provide relatively complex structured output. With that said, DV was designed with the intent to be configurable or to be enhanced to use almost any other embedding mechanism or LLM (including LLMs incapable of providing more structured output), though those capabilities have been minimally tested at most.

DV can be configured to query a PostgreSQL database to populate its tables with one or more terminologies, and it can be configured to query a database to generate code subsets from the terminologies (such as "diagnosis" or "medication" subsets). When populating content (e.g., from the UMLS Metathesaurus[10]), it can be configured to automatically generate and store embeddings for terminology strings (text), so that these computationally expensive operations need only be performed once. DV can store multiple embeddings per string, where each embedding was generated by a different model. One potential advantage of SentenceTransformers models is that they can generate a "CLS" token (a special kind of embedding) for each phrase processed. This token is designed to represent the overall meaning of the phrase.[11] Alternative methods for representing a multi-word phrase with a single embedding might be to use "pooling" such as mean pooling, where the embeddings of individual words are averaged together into a single embedding. In the anecdotal experience of this author during DV development, CLS tokens seemed to provide better performance at some tasks compared to mean pooling in some contexts. Therefore, DV will often use CLS tokens for certain tasks. In this work, where CLS tokens were used, please note that other forms of tokens (e.g., mean-pooled or max-pooled) could have been used instead, although some code changes might be required for that.

DV can then be configured to generate relationships for all items in a terminology, a code set, or returned by a database query. It can do this by executing a database query (e.g., populating its knowledge graph from other knowledge graphs stored in a database, by using SQL query-based manipulations of database data to infer relationships, etc.), or by using queries of an LLM. This work focuses on populating the DV knowledge base using LLM queries. To do so, for the desired set of terms (e.g., all terms in a code set), it will iterate through the set and, one at a time, substitute each item into a configured LLM prompt and submit the resulting prompt to the LLM. For example, for a set of diagnosis terms, DV might be configured with a prompt such as "What are the medications that treat <<<concept>>>?" DV would then iterate through the list of diagnosis terms, one-by-one substituting "<<<concept>>>" with a term, submitting the resulting prompt to an LLM, and then processing the response. DV will then attempt to parse each LLM response into a structured format, then store the results in its database. The set of relationships in its database are the DV knowledge graph. The knowledge graph is stored as a "triple store," a database table with each row containing a relationship “triple.” Each relationship triple is composed of 3 key elements (database fields): the “subject” (in this case, a terminology concept entry), the “predicate” (i.e., the type of relationship), and the “object” (in this case, an item from the LLM’s response). For example, since one possible complication of myocardial infarction might be congestive heart failure, an example triple might be:

- **Subject**: the SNOMED-CT code for “myocardial infarction”
- **Predicate**: “has_complication_of”
- **Object**: “congestive heart failure”

gpt-4o-mini costs less for a set of similar transactions if only the latter part of the prompt changed with each query, so configured and auto-generated prompts were designed, as much as possible, to have “<<<concept>>>” at or near the end of the prompt.

Where the LLM offers compatible functionality, DV attempts to keep track of the total number of tokens consumed during a “run” (an instance of processing a series of terms), and to automatically kill the queries when a dollar limit has been exceeded. This has some caveats (e.g., the cost per token must be correctly configured in the system) and some limitations (e.g., if a set is processed in multiple runs/executions of the system, the total cost across all runs is not tracked or managed). Despite these limitations, DV may help to manage costs and avoid cost overruns when working with a commercial (i.e., pay-per-use) LLM service, beyond any related functionality that may be offered by the LLM service itself.

## ERRONEOUS, OFF-TOPIC, AND DEPENDENT RESPONSES

An LLM may provide more accurate responses if allowed to reason over the prompt prior to providing a final answer.[12] To allow for this, DV can be configured to identify response elements as “no write” to hold the LLM reasoning content while denoting that such content need not be stored in the knowledge graph.

At times the author noticed that one response would depend on others, while at other times, a better response to one query could be achieved if the LLM provided it along with the response to another query. For example, if querying for the proportion of patients with a disease having a mild, moderate, and severe course, the LLM seemed much more likely to provide responses that sum to 100%, and to provide more accurate responses for each level of severity, if asked to provide all proportions in a single prompt as opposed to being asked for each as a separate prompt. To support this, DV has functionality to facilitate and handle multiple relationship queries in a single prompt.

To further address potentially erroneous responses, DV offers an “are you sure?” functionality, which allows the user to configure DV to automatically re-query the LLM one or more times. However, at least one paper noted that a single re-query of an LLM asking if it is sure of a response (with another final re-query if the first re-query did not actually provide an answer) may lead to less accurate responses (that paper even gave the same “are you sure” name to the functionality).[13] Therefore, DV allows zero, one, or more such queries, and offers a few methods to finalize the relationship:

- **Vote**: If the prompt expressly requests that a particular response item should only be a value from a list of allowed values (i.e., the response is categorical), then DV can use “voting” to finalize the object of the relationship. The most common response becomes the object of the relationship.

Although not necessary to have done so, DV allows for a different configuration for the special case of a Boolean-like categorical response limited to a value of 0 or 1.

- **Average**: If the prompt expressly requests that a particular response item should be numeric, then DV can average the responses to generate a final value as the object of the relationship.
- **Sum**: If the prompt expressly requests that a particular response item should be numeric, then DV can sum the responses to generate a final value as the object of the relationship.

## FREE-TEXT RESPONSES

In the process of developing DV, the author observed that mapping LLM output back to terminology codes using vector comparisons (such as cosine distance) seemed insufficiently accurate. DV can generate subject summary vectors (a.k.a. code summary vectors) that pool (e.g., via mean pooling) the vectors (e.g., the CLS tokens) of all strings associated with a terminology concept or code into a single string. Pooling was used for aggregating CLS tokens instead of concatenating the underlying strings into a single phrase and generating a CLS token from it because pooling (at least anecdotally) appeared to lead to better performing embeddings in that context. However, whether using a UMLS-asserted preferred string for a concept ("main subject string"), or a pooled embedding representing the aggregate of the strings (mean-pooled or max-pooled), the performance of the resulting embeddings for mapping an LLM text response to a corresponding terminology code seemed inadequate. The investigator noted anecdotally that the CLS tokens often seemed to have smaller cosine distances when the phrases contained the same words than when the phrases differed in their words but had the same or similar apparent meanings. This seemed especially true when phrase had more words than the other.

To address, this, DV was coded to be able to automatically re-query an LLM to generate multiple versions of a string, and to do so in one or more particular "styles." In DV, these are called "expansion strings". An example of a style might be to express something simply, so that alternative versions of "fracture of an unspecified upper extremity digit" might be "broken finger" or "finger fracture." DV can do this for the string(s) represented by a code, and for a string returned as part of an LLM's response to a prompt. Thus, strings having the same meaning but different "styles" (e.g., a string from the ICD-10 terminology and the string returned by an LLM) may match more closely when converted into the same style. Through this mechanism, free-text LLM responses may be better matched to structured terminology codes.

Various strategies can then be used to match an LLM response to a terminology code. For example, the multiple strings available for the original string or code can be aggregated (e.g., via mean-pooling) and then the resulting mean-pooled vectors can be compared to one another. As another example, the closest matching mean-pooled vectors across two compared sets may be used.

Anecdotally, the use of expansion sets seemed to improve the accuracy of matching an LLM's string output to an appropriate code in a terminology.

With the aim of reducing costs and processing time, DV saves every expansion string in its database, and will only re-query the LLM to expand a string if it hasn't already previously gotten an expansion set for the string in the past using this LLM.

For every triple (every subject, predicate, object relationship), DV can be configured to expand its "subject string" into an "expanded subject string set" and its "object string" into an "expanded object string set" using one or more styles. DV will generate and store a vector representation (embedding) of each aforementioned string, and to generate "summary vectors. See Figure 1 for a visual representation of many of these key strings and vectors. Note that:

- "Subject Expansion Summary Vector" has not been implemented yet as of the time of this writing.
- In many or all cases, for items labeled as "string vectors," more than one vector is created (e.g., mean pooling vector, max pooling vector, a CLS vector, etc.).
- The boxes labeled as "mean-poolings" currently mean pool the relevant CLS tokens.
- "Object expansion string vector" is calculated when creating the Object expansion summary vectors but is not currently stored by the system. That may be a potential future enhancement.

Figure 1

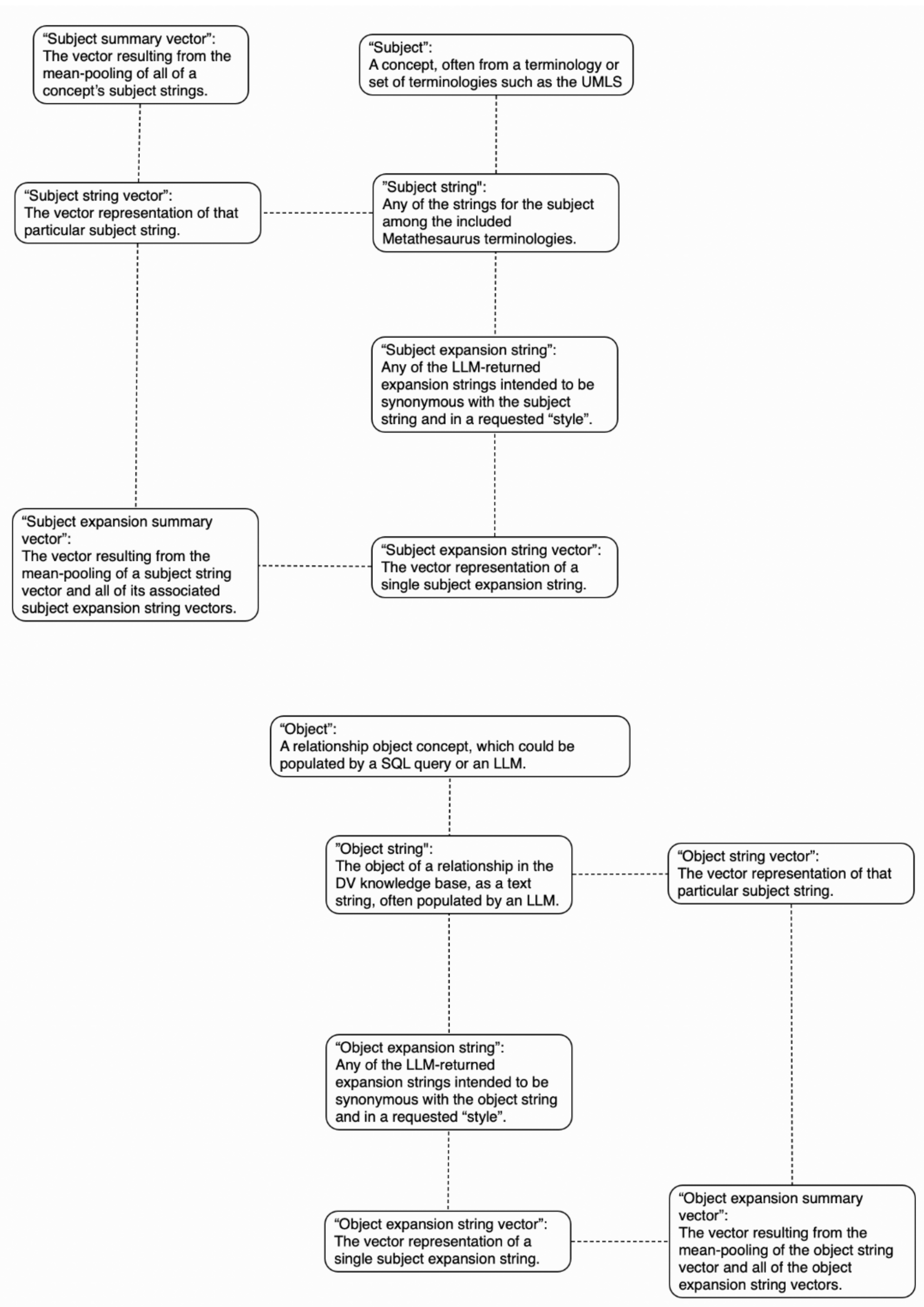

Although various strategies could be used to match a free-text LLM response to a terminology code, such mapping seemed like it should be a first-class function of the system. Therefore, DV has specific functionality for this, matching object string vectors having a user-selected “style” to the closest matching (smallest cosine distance) subject vector having that same style within a selected code set. The system allows the user to select one or more subject vector types and one or more object vector types for the comparison. To see this in mathematical terms, see Figure 2 (generated by ChatGPT [14] based on my description of the code and query logic, then its LaTeX code was pasted into the Overleaf online LaTeX editor [15], from which a screenshot was used for the image).

Figure 2

## Definitions

Let $x$ be the input string.

Let $\mathcal{C}$ be the set of codes, where each code has associated vectors.

Let

$$\mathbf{V}_x = \{\mathbf{v}_x^{(1)}, \ldots, \mathbf{v}_x^{(m)}\} \subset \mathbb{R}^d$$

be the set of vectors associated with string $x$.

Let

$$\mathbf{V}_c = \{\mathbf{v}_c^{(1)}, \ldots, \mathbf{v}_c^{(n)}\} \subset \mathbb{R}^d$$

be the set of vectors associated with code $c \in \mathcal{C}$.

Let the cosine distance between two vectors be defined as

$$\text{cos_dist}(\mathbf{u}, \mathbf{v}) = 1 - \frac{\mathbf{u} \cdot \mathbf{v}}{\|\mathbf{u}\| \|\mathbf{v}\|}$$

Let $z \in [0, 2]$ be the maximum allowable cosine distance.

## Formulas

The distance between string $x$ and code $c$ is defined as the minimum cosine distance between any pair of vectors associated with them:

$$d(x, c) = \min_{\mathbf{v}_x \in \mathbf{V}_x,\, \mathbf{v}_c \in \mathbf{V}_c} \text{cos_dist}(\mathbf{v}_x, \mathbf{v}_c)$$

The best matching code $c^*$ is the one minimizing this distance, subject to the threshold $z$:

$$c^* = \arg\min_{c \in \mathcal{C}} d(x, c) \quad \text{such that} \quad d(x, c^*) < z$$

Since such computations, even with vector indexing, appeared to be relatively computationally expensive, the system will store the matches in the database for fast, low-cost, low computation retrieval afterward. Since the matching will not always be correct, the system will store the top four matches by default, in the hope that the most correct match will be in the top four. The instance of the class that performs this operation can be configured to store a different value for the top "n" matches other than four.

When a categorical response was desired (i.e., a response within a specifically constrained list of possible responses), it was noted that the free-text response by an LLM may not always perfectly match an item in the provided list of acceptable responses. The LLM may sometimes paraphrase or make other minor changes, leading to an inexact match to an item in the list of requested potential responses. For example, if one option in the list was "fracture", an LLM might return "fractures" or "bony injury" or "broken bone". However, it was also noted that gpt-4o-mini seemed to more reliably return a letter or short word denoting one of the desired categorical responses. For example, the model might be provided with the following:

Answer only with one of the appropriate letter values that follow:

a: Headache with movement
b: Fracture of femur
c: Fever of unknown origin

In a case akin to the above, the model seemed much more likely to accurately return only the letter corresponding to its answer than the exact desired answer itself. Given this, DV allows the user to configure a JSON dictionary to go in the prompt, with the dictionary having the letter, number, word, or term as the key and the desired categorical response as the value. DV will format the dictionary into the prompt and then convert responses from the LLM into to their categorical string value for storage.

Since, as noted below, DV may also ask for a number serving as a measure of the specificity of the provided response, the LLM occasionally seemed to confuse the two numbers. For this reason, the author changed to using letters or short terms instead of numbers as the dictionary keys for categorical responses.

## OVERLY GENERAL RESPONSES

DV allows the user to configure either of several mechanisms to determine if an LLM response item was overly general, and if so, to automatically re-query the LLM to replace that item with a more specific set of items. To avoid causing ambiguity by using a term for this issue that could have other meanings (e.g., specificity), the author created the term “beceptivity” to indicate a value on a scale ranging from most general to most specific. Higher beceptivity indicates a concept that is more specific and/or detailed whereas a lower beceptivity indicates a concept that is more general and/or vague. Users may configure virtually any beceptivity scale provided that the lowest end of the scale is zero and higher values signify greater beceptivity (less general, more specific). If DV has been configured with a minimum required beceptivity, DV will automatically re-query for “more beceptive” responses. These responses will replace the inadequately beceptive response.

DV can be optionally configured to use one of 3 primary methods to check each response for adequate beceptivity:

1. Users can configure DV prompts that will return the LLM’s numeric assessment of the beceptivity along with each response item. This has the advantage of reducing the number of queries posed to the LLM with the goal of reducing cost and processing time. However, it was anecdotally noted in this work that the LLM occasionally had difficulty returning such a complex structure and returned poor beceptivity values more often than with other approaches.
2. Users can configure DV to re-query the LLM to provide a beceptivity value for each returned response item. To save time and costs, DV will store the returned beceptivity so that the LLM does not need to be re-queried for its beceptivity if the same string is returned in another response. However, the more beceptive responses returned on a re-query are not stored and reused with new queries using different concepts. For example, consider a prompt of “what are the medications used to treat <<<concept>>>?” Imagine that two of the concepts used to replace “<<<concept>>>” in the prompt are “urinary tract infection” and “staph skin infection.” Finally, imagine that the LLM returns an answer of “antibiotics”. The more beceptive values for “antibiotics” will differ when the concept in question is “urinary tract infection” than when the concept is “staph skin infection”. For this reason, although the beceptivity magnitude of “antibiotics” is stored and reused, the more beceptive versions of the term are not.

3. Users can configure DV to query a database for a string’s beceptivity. For example, one might query UMLS Metathesaurus data to assess a string’s beceptivity based on its location on a hierarchical tree.

## STRUCTURE OF THE RESPONSE

Based on the configuration parameters, DV will automatically generate a part of the prompt that instructs the LLM on how to format its response. The formatting instructions differ depending on whether the LLM can return a relatively complex JSON object in its response. The returned response, whether JSON or another format (e.g., a series of pipe-delimited text snippets), gets parsed into a Python object by DV using Python libraries and/or custom Python code. So far, only the relatively complex JSON requested of, and returned by gpt-4o-mini has been reasonably (or at all) tested.

The structured response facilitates the downstream handling of responses (such as re-querying the LLM when an object string is inadequately beceptive or storing each resulting predicate and object string for a subject in the DV knowledge graph triple store [database table]).

## USER EXPERIENCE

Although the system can be configured using configuration files or editing endpoint code, it seemed clear that the greatest uptake and fastest and easiest use of the system would require a graphical user interface (GUI). Therefore, a web browser-based GUI was designed leveraging standard HTML, JavaScript (JS), the Preact version 10.26 JS library [16], and Cascading Style Sheets, with server-based endpoint management and some other functions handled by Flask [17], along with other libraries and tools.

Through the GUI, the user can:

- Populate the DV database with a terminology of codes and associated strings by querying another table in the database containing the source material. DV will also automatically create string vectors for all strings associated with the codes and code summary vectors.
- Create a code set, which is a subset of codes from a terminology (such as the subset of codes in the terminology that represent medications). DV can be configured to automatically create expansion strings in a “style” for each of the strings in the code set.
- Generate relationships to codes in a code set (see Figure 3 for a screenshot of part of that page) by querying an LLM using prompts. DV will automatically create string vectors for all strings associated with the codes, and DV can be configured to automatically create expansion strings in a “style” for each of the object strings, as well as string summary vectors.
- Match relationship object strings to the closest matching code(s) in a code set (e.g., using cosine distance).
- Create custom tables using a query, for example, one that combines DV code sets, DV relationships, and data from non-DV data sources. This materialization of the query may be helpful for significantly enhancing performance of complex queries that may be needed frequently. It also provides a mechanism to store and reprocess the query and its materialization.

# EXAMPLES

During DV development and subsequent exploration and functional testing, the MRCONSO table of a subsetted UMLS Metathesaurus was imported into the DV database, with the UMLS Concept Unique Identifiers (CUIs) as the code and all associated strings for each code also imported and associated with their relevant CUIs. Several code subsets were generated, including the SNOMED-CT Core Problem List, the ICD-10 CM codes, a code subset containing both of those codes, the RxNorm medication brand names, and the RxNorm medication ingredients.

Several relationships were generated for relevant code subsets, such as test findings, causes, differential diagnoses, physical findings, and complications of clinical problems. For at least one of these, code mappings were generated for the relationship objects. On informal review by the author, the generated relationships and mappings of object strings to codes seemed valid far more often than not, though formal analysis and validation is anticipated in follow-up work. Since not every LLM-generated relationship or other output of DV will be perfect (e.g., incorrect, overstated, overly general, incorrect code mappings, etc.) despite the many features built into the system to minimize this, the intended use cases for the generated knowledge graphs are those designed for minimal risk of harm in the face of erroneous relationships, perhaps for example, in expert-in-the-loop workflows involving quality of care reviews and analyses.

Figure 3

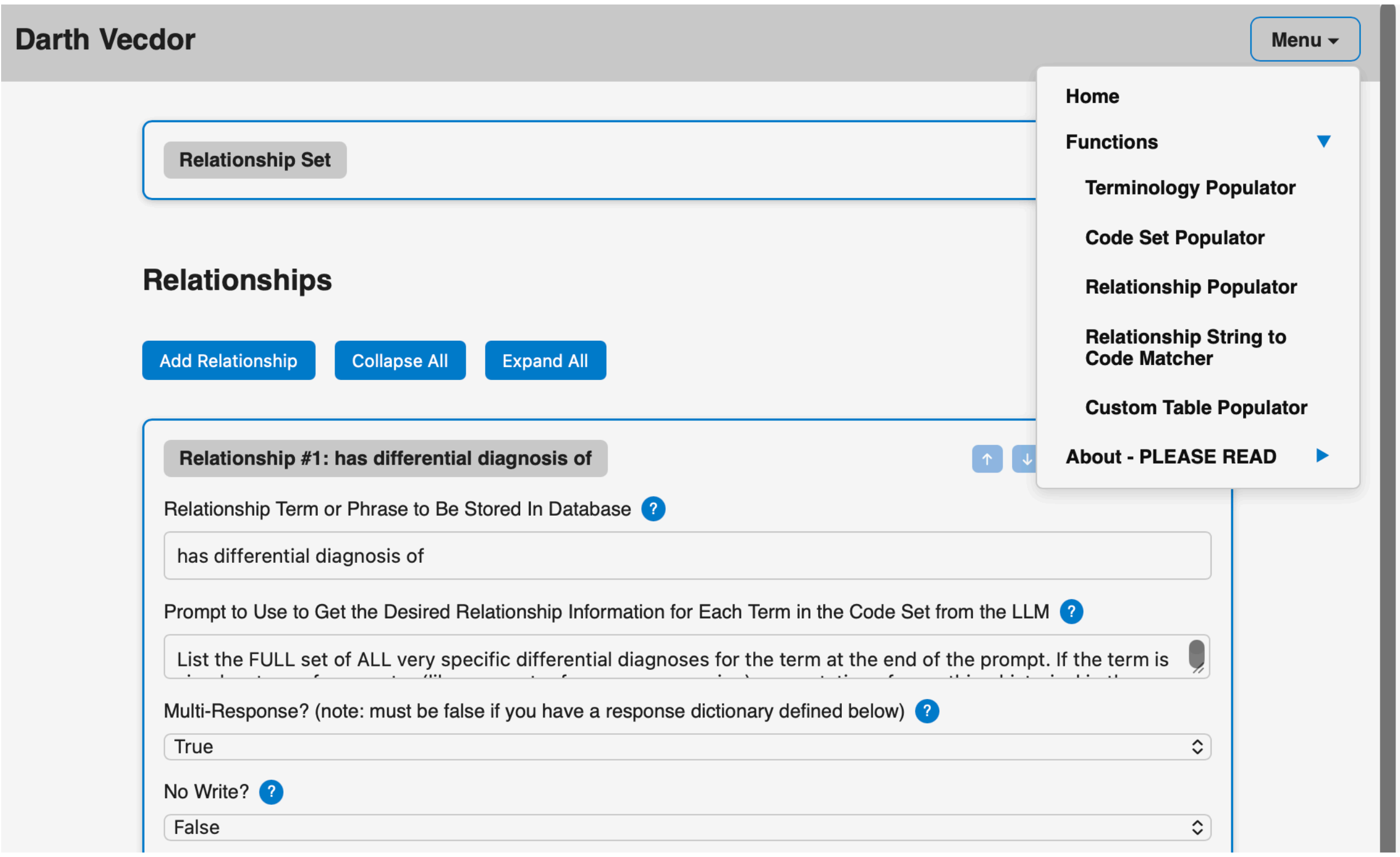

# DISCUSSION

## RELATED WORK

Others have used knowledge bases to perform Retrieval Augmented Generation (RAG) with LLM queries [18], [19], [20] and to fine-tune LLMs for a specific purpose,[21] and have used LLMs to extract content from medical documents (such as journal articles or guidelines) to create knowledge bases.[22], [23], [24], [25] Even closer to this work, others have seemed to describe the population or potential population of a structured data knowledge base from iterative queries of an already-trained LLM (i.e., an LLM not specifically trained for that work).[26], [27], [28] However, a search of PubMed on the quoted terms "large language model" and "knowledge base" and a PubMed search on the quoted terms "large language model" and "knowledge graph" [29], [30] returned no articles that appeared to describe a general-purpose framework akin to DV and with DV's many key features and functions. In that search, articles were assessed based on the content of their abstract and, when freely available to the author determined potentially relevant, their full text. Although the full text was not available to the author for all potentially relevant articles, all such cases seemed unlikely to have all of DV's key features and functions based on the abstract content.

Although this paper describes the use of DV with an LLM not specifically trained or fine-tuned for any particular downstream task, the system was designed with configurability as well as a "plug-in" architecture that should allow DV to work with virtually any fine-tuned or purpose built LLMs that offer an API for queries.

## LIMITATIONS

Although this paper's content seems to imply that DV might be used or be useful for medical or healthcare (or for any other purpose), it may not be suitable for many such purposes, or even for any purpose, and may even be dangerous and harmful. DV is distributed on an "AS IS" BASIS, WITHOUT WARRANTIES OR CONDITIONS OF ANY KIND, either express or implied.

The choice of whether and how to use DV is entirely up to the user, all use is at the user's own risk, and there are no warranties or conditions of any kind. Users should read and understand the DV license, and nothing here supersedes those terms and conditions.

In addition to all the caveats in, and elements of, the license for use of Darth Vecdor, in this document, and throughout the codebase, great care should be exercised in using Darth Vecdor or any of its outputs. DV should only be used by those with the appropriate expertise and experience needed to determine the proper use and implementation of DV, and to differentiate correct from incorrect responses.

DV generally does not contain security or privacy features or even basic security functionality, capabilities, or even necessarily best practices. It is up to the user to configure DV for secure and private use to the extent appropriate and possible. None of this should be assumed present by default (and much of it definitely is not). Anything that appears to involve security or privacy should not be trusted. Security and privacy are entirely up to the user.

Not all elements of DV have been tested, and many have only been tested cursorily. DV may have bugs and other very serious code, design, database, and other issues. Some of those may already be known by the author or others, and some may not yet be known. Only those with the appropriate expertise should consider the possible implementation and use of DV, and the risks of doing so must be carefully considered.

Some features, capabilities, or other characteristics of DV described here may have changed or been removed since the time of this writing, although removal of features seems likely minimal, if any. This paper may contain errors and omissions.

Neither DV nor its outputs have been formally validated. All validation and use are the responsibility of the users and/or future researchers. This listing of caveats and warnings should not be considered a complete listing. Validation of a platform's effectiveness (as opposed to verification that it performs its functions as designed) can be challenging. Virtually any validation of the utility of DV as a whole or the utility of particular DV outputs would likely be specific to the use-case, the underlying LLM used, the selected embedding-generation mechanism, and the DV prompts and other configuration parameters that were applied. A failed validation of DV outputs might succeed if prompts were modified, DV configurations were modified, or if a different and more capable LLM were used. For this reason, a positive validation can demonstrate the feasibility of DV for a particular use, but it may be that no generalizable conclusion can be drawn from a failed validation of DV outputs other than that the specific set of prompts, configurations, LLMs, and embedders did not meet the performance target.

## CONCLUSION

Knowledge graphs are a useful and important tool in informatics but obtaining a comprehensive knowledge graph for healthcare may be challenging. For example, although the UMLS MRREL table has many millions of relationships and has substantial value for many uses, this author has found many relationships inadequately comprehensive for many healthcare-related needs. DV offers a potential path to fill these gaps and to create new and novel knowledge graphs using the information encoded in LLMs. As LLMs and associated technologies improve, this author anticipates that the knowledge graphs they can produce will be of increasingly higher and more consistent quality and utility. However, even in their current state, it seems likely that the LLMs may produce knowledge graphs that have utility for selected use cases. Identifying these use cases and assessing the utility of DV-produced knowledge graphs for those use cases represents an important area for future research.

## SUPPLEMENTAL MATERIAL

Darth Vecdor's source code is available at: https://github.com/jonhandlermd/darth_vecdor

## FUNDING

The development of Darth Vecdor was sponsored, at least in part, by Keylog Solutions LLC.

## ACKNOWLEDGEMENTS AND ADDITIONAL INFORMATION

Darth Vecdor was entirely invented, designed, created, developed and built by Dr. Jonathan Handler on his own and/or in his role with Keylog Solutions LLC (with the help of large language model[s] and many other resources), completely outside of and unrelated to his employment with OSF HealthCare. The author wishes to thank Dr. John Vozenilek, of OSF HealthCare, for his review of this paper.

## DECLARATION OF INTERESTS

Dr. Jonathan Handler is chief executive officer and president of Keylog Solutions LLC and a shareholder of Keylog Solutions LLC; has received research-related funding from Pfizer; is a shareholder in other healthcare companies including Whispersom Corporation, EmOpti LLC, HealthLab LLC, and Baxter Healthcare; serves in an advisory role to Whispersom Corporation, EmOpti LLC, and HealthLab LLC; has various patents that have been granted or are pending; and related to activities for the respective institutions has received stipend, food, and/or travel from University of Michigan and the American Medical Association. Dr. Handler is an employee of OSF HealthCare.

https://pubmed.ncbi.nlm.nih.gov/?term=%22knowledge+graph%22+%22large+language+model%22&sort=date